\documentclass{article} 
\usepackage{times}
\usepackage{url}
\usepackage{graphicx}
\usepackage{amsmath}

\title{A Survey: Time Travel in Deep Learning Space: \\
An Introduction to Deep Learning Models\\
and How Deep Learning Models Evolved from the Initial Ideas}

\author{
Haohan Wang and Bhiksha Raj\\
\\
Language Technologies Institute\\
School of Computer Science\\
Carnegie Mellon University\\
Pittsburgh, PA 15213 \\
\texttt{\{haohanw, bhiksha\}@cs.cmu.edu} \\
}

\begin{document}

\date{}
\maketitle

\begin{abstract}
This report will show the history of deep learning evolves. It will trace back as far as the initial belief of connectionism modelling of brain, and come back to look at its early stage realization: neural networks. With the background of neural network, we will gradually introduce how convolutional neural network, as a representative of deep discriminative models, is developed from neural networks, together with many practical techniques that can help in optimization of neural networks. 
On the other hand, we will also trace back to see the evolution history of deep generative models, to see how researchers balance the representation power and computation complexity to reach Restricted Boltzmann Machine and eventually reach Deep Belief Nets. 
Further, we will also look into the development history of modelling time series data with neural networks. We start with Time Delay Neural Networks and move further to currently famous model named Recurrent Neural Network and its extension Long Short Term Memory. We will also briefly look into how to construct deep recurrent neural networks.  
Finally, we will conclude this report with some interesting open-ended questions of deep neural networks.

\end{abstract}

\newpage
\tableofcontents
\newpage

\section{Introduction}
This report is an introduction of deep learning models. Different from other deep learning tutorial, this report aims to draw a very comprehensive and yet brief picture about how these deep learning models are developed. 
It is comprehensive because is covers a lot of initial ideas about network models, like basic Neural Network, Self Oraganizing Map, Time Delay Neural Network. 
Yet it is brief because we cannot introduce everything about deep learning models due to the fast development of this field.

In this report, we will first trace back to the very beginning of how researchers start to model brain. 
With a belief of connectionism, a premier notion of neural networks is formed. 
Further, artificial neural networks are introduced with the successful design of perceptrons and backpropagation algorithm and start to play a dominating role in machine learning area.

With the background of early stage work, the Section 3 is devoted to some famous discriminative deep learning models. 
From the necessity of depth of models, we first move to stacked autoencoders and then introduce convolutional neural networks and some of its infinitely many optimizing techniques. Most of these techniques could be generalized to other deep learning models.

Section 4 is devoted to generative models. 
Again, we travel through a long journey from the early stage work like self organizing map and Hopfield network, to the middle age Boltzmann Machine and Restricted Boltzmann Machine, and to nowadays Deep Belief Nets, which is a natural extension of earlier work.

Then, Section 5 is a collection of neural network models for time series data, or generally sequential data. 
As a pioneer, Time Delay Neural Network is introduced first, followed by Recurrent Neural Network and the famous model called Long Short Term Memory (LSTM). 
Then, an intuition of how these models could be benefited from deep representation learning is introduced at the end of that section.

Last but not least, Section 6 is about some other interesting topics of deep learning and some topics that are yet to be fully explored when this report is prepared. Here, we will actually show some validation for the claim that deep learning is the best model so far. However, we will also see some flaws of the deep learning, that may possibility hinder its further development. 

\section{From the Very Beginning to Neural Network}

\subsection{Connectionism and Early Stage of Artificial Neural Network}
\label{post:1}
This very first section is devoted to the very initial stage of Artificial Neural Network. It is built on people's one point of view of understanding human mental phenomena: connectionism. 
The remaining is about the pioneer work of neural network modeling, how the information is passed and stored in the network and the early mathematics model. 
These works may not be directly useful for understanding neural network. However, we believe this whole process will be helpful for the researchers who are interested in pushing the deep learning work forward. 
\subsubsection{Connectionism}
It is a long way for researchers and philosophers to go in understanding how human brain and neural system function. 
Finally, they end up in two different, yet both interesting tracks: connectionism \cite{fodor1988connectionism} and computationalism\cite{dietrich1990computationalism}. 
Needless to mention that which path is more promising or which path is the subset of the other, here we only focus on the connectionism points of view, where people believe that human intellectual abilities can be explained with a network of very simple units (neuron). 

\subsubsection{Early Stage Pioneers}
\begin{figure}[h]
\center
\includegraphics[scale=0.15]{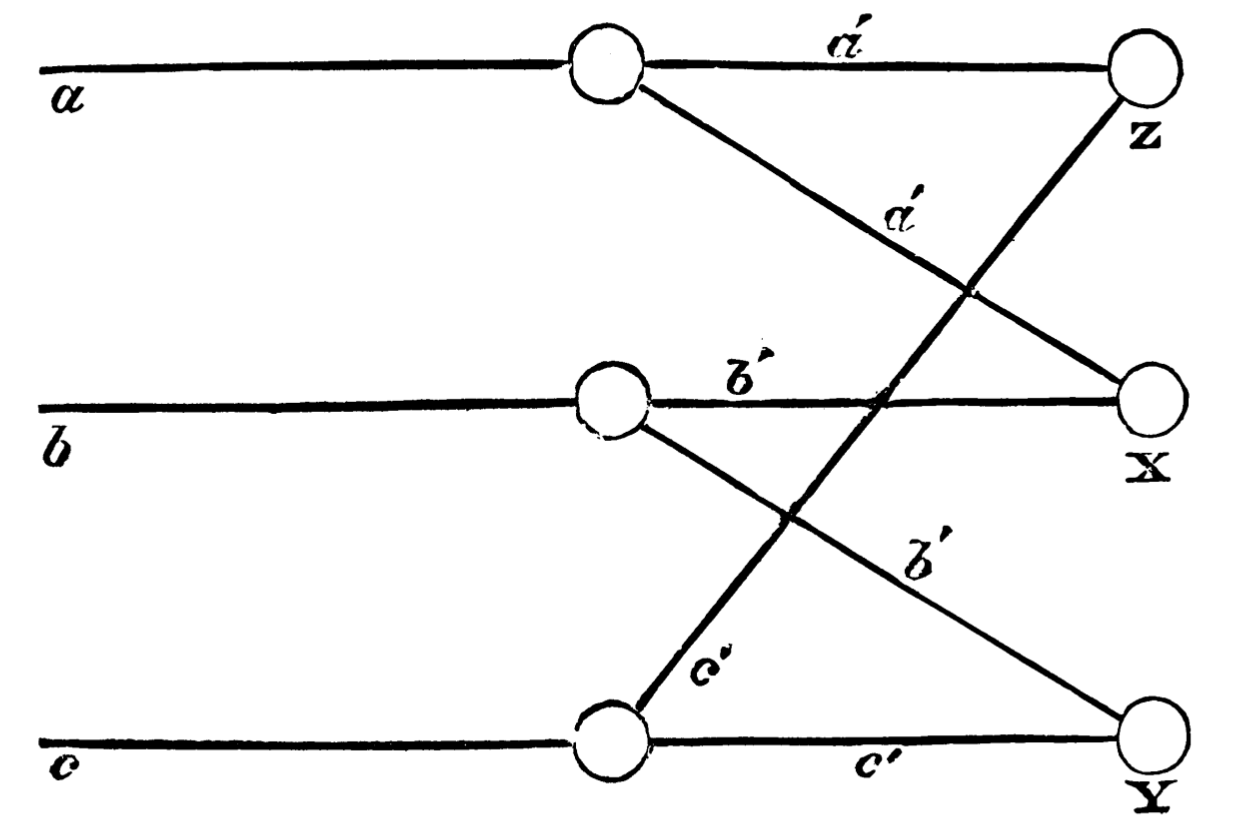}
\caption{An example of Bain's neural groupings structure. (Figure from \cite{bain1873mind})}
\label{fig:bain}
\end{figure}
Relevant work started in 19th century. To make it easier to understand, it starts out as a philosophy question, rather than a computer science problem or a biology problem.

Alexander Bain first mentioned the connectionism ideas in \cite{bain1873mind}. 
Even earlier, he offered an interpretation of mental phenomena within an association framework. Then, in his book, he argues a brand new point of view stating memory is a model capable of "putting required stuff together" rather than "storing everything beforehand". 
He named this memory structure as "neural groupings", as a predecessor of "neural network" in our world, and further explained that under his view, connected nerve fibers are channeled to different parts of the network under different simulations. 

As a very simple example, showed in Figure~\ref{fig:bain}. $X$, $Y$, $Z$ is triggered by different combinations of simulations from $a$, $b$ or $c$.

Bain's work is recognized as the earliest work of discovering neural networks of human brain, however, Bain was not convinced by himself and argued that this structure shows no practical value. 

Donald O. Hebb was influenced by Bain. 
In 1949, he stated the famous rule: "Cells that fire together, wire together" \cite{hebb1949organization}, which emphasized on the activation behavior of co-fired cells. 
This theory attempts to explain neural groupings theory and provides a biological basis for learning methods in the rehabilitation of memory. 
In a more formal way, it says that when an axon of cell A is near enough to excite cell B and repeatedly or persistently takes part in firing it, some growth process or metabolic change takes place in one or both cells such that A’s efficiency as one of the cells firing B is increased. Mathematically, it could be written as: 
\begin{align}
\Delta \omega_i = \eta x_i y
\end{align}
for the changes of synaptic weights.

The weight of neurons/connection units get updated with this learning rule, indicating that the connection of two signals is magnified every time when they appear together, which is consistent with our daily experience as human. 

\subsubsection{Perceptron in Neural Network}
As these pioneers successfully brought the connectionism and neural groupings into the world, other researchers started to work to complete the whole theory. 
One important completion is Frank Rosenblatt's thoughts on how the information is stored and transferred in neural network.

In \cite{rosenblatt1958perceptron}, he raised three questions regarding perceptrons: how perceptrons can help to 1)detect information, 2)store information and 3)recognize something with the information.

\begin{figure}[h]
\center
\includegraphics[scale=0.175]{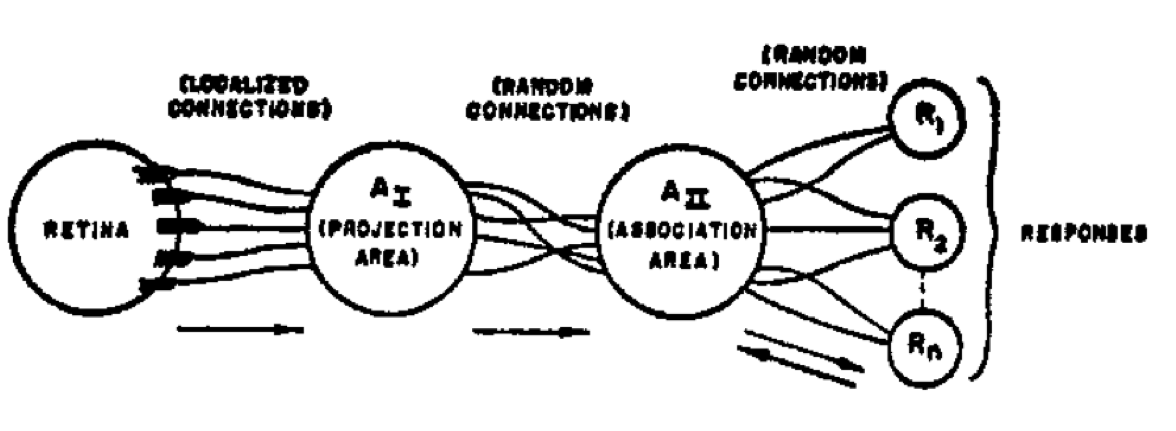}
\caption{An illustration of Rosenblatt's model of perceptron (Figure from \cite{rosenblatt1958perceptron})}
\label{fig:frank}
\end{figure}

The first question is quite clear. 
The answer to it belongs to some work within the sensing field, thus this is irrelevant with this report.
As for the following two questions, inspired by the connestionism, Rosenblatt believes that simple logic and boolean algebra cannot explain the phenomenon very well and he preferred probability theory. 
He made several assumptions to the model accounting for randomness, plasticity and similarity and he proposed a model of perceptron with S-points (as input), A-units (as activation function in model neural networks) and R-cells (as output). 
The model is showed in Figure~\ref{fig:frank}.

\subsection{Perceptron as a Principal Component Analyzer}
\label{post:2}
After some historical points of views that built the foundation of neural network and deep learning previously, we can talk about computer science stuff now. 
Here, we will talk about the most basic unit of a neural network: the neurons. First, we will introduce perceptron, as the most basic neuron, then introduce the reasoning that a perceptron is basically a principal component analyzer.

\subsubsection{Perceptron}
\begin{figure}[h]
\center
\includegraphics[scale=0.5]{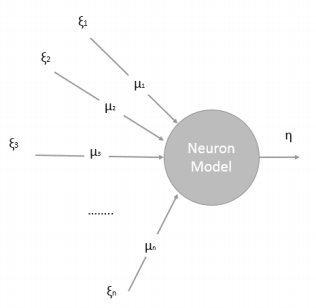}
\caption{A linear perceptron}
\label{fig:perp}
\end{figure}

A linear perceptron builds from the assumption that the data labels we are interested in can be (approximately) represented by linear combination of the features. Thus, it can only make decision based on linear boundaries. A linear perceptron is shown as Figure~\ref{fig:perp}.

In Figure~\ref{fig:perp}, vector $\xi$ stand for the features of a data, $\mu$ are the weights corresponding to each feature, a perceptron takes the linear combination of features and make decision by comparing the weighted sum to a threshold. 
The weights and threshold are the parameters of this perceptron, learnt through machine learning.   
How do these weights get learnt? 
A simple weight update function is designed according to Hebbian learning rule, following the idea cells that fire together, wire together. Refer to Section~\ref{post:1} for details.

With the basic understanding of linear perceptron, other non-linear neurons can be easily generalized with other forms of activation functions replacing linear thresholds. 
A sigmoid function is one of the most famous ones. 
When other functions are applied, simple updating rule may not be the best choice. 
Gradient descent is frequently adopted to update the weights. 

\subsubsection{Perceptron as a Principal Component Analyzer}
Now, let's focusing on an interesting theory that when a perceptron is trained, the weights are updated in such a way that principal components of the data are extracted. In other words, a perceptron is actually doing the job of a Principal Component Analyzer \cite{oja1982simplified}.

Before we start, let's make sure we are on the same page for PCA \cite{person1901lines}. PCA is used to describe the data set with a series of principal components, where each principal component maximizes its variance and orthogonal to each other. There are three points of views to see PCA:

\begin{enumerate}
\item It maximizes the variance of principal component, as showed in following:
\begin{align*}
var & = (Xu)^T(Xu)\\
u &= \arg\!\max_{u^Tu=I}(var) = (Xu)^T(Xu) = u^TX^TXu\\
u &= \arg\!\max_{u^Tu=I}u^TCu
\end{align*}
\item It minimizes the reconstruction error, as showed in following: 
\begin{align*}
E &= \sum_i||x_i - (x_iu)u||^2\\
&=\sum_i (x_i^Tx_i)-2(x_iu)x_i^Tu + (x_iu)^2u^Tu\\
&=\sum_i (x_i^Tx_i) - 2(x_iu)^2 + (x_iu)^2\\
u &= \arg\!\min_{u^u=I}E = \arg\!\min_{u^u=I}-\sum_i(x_iu)^2=\arg\!\max_{u^Tu=I}u^TCu
\end{align*}
\item Components are eigenvectors and the first component is the eigenvector corresponding to largest eigenvalues of covariance matrix. as showed in following: 
\begin{align*}
L &= u^TCu - \lambda (u^Tu-I)\\
\dfrac{\partial L}{\partial u} &= 2Cu - 2\lambda u = 0\\
Cu &= \lambda u
\end{align*}
\end{enumerate}
It is interesting and quite trivial to see that these three views are equivalent. 

Now Let's start from the perceptron in Figure~\ref{fig:perp}. 
All the functions above are summarized into the middle circle.

With these notations, a decision can be made through the following function:
\begin{align}
\eta = \sum_i^n\mu_i\xi_i
\end{align}
and according to hebbian hypothesis, the weights of this perceptron can be updated through the following equation:
\begin{align}
\mu_i(t+1) = \dfrac{\mu_i(t)+\gamma\eta(t)\xi_i(t)}
{\sqrt{\sum_i(\mu_i(t)+\gamma \eta (t)\xi_i(t))^2}}
\end{align}
where it is the basic hebbian learning rule with a normalization term of L2 norm of the equation.

Then, a new law is introduced \cite{oja1985stochastic}, where we assume the gain coefficient is small and don't consider second terms, like following: 
\begin{align}
\mu_i(t+1) = \mu_i(t)+\gamma\eta(t)(\xi_i(t)-\eta(t)\mu_i(t)) + O(\gamma^2)
\end{align}
With this function, the growth of weights is controlled with internal feedback.

Now, we look into the problem with a matrix point of view:

\begin{align}
\mu(t+1) = \dfrac{\mu(t)+\gamma\eta(t)\xi(t)}{||\mu(t)+\gamma \eta (t)\xi(t)||_2}
\end{align}

same as above, we now have: 
\begin{align}
\mu(t+1) &= \mu(t) +\gamma\eta(t)(\xi(t)-\eta(t)\mu(t)) + O(\gamma^2)\\
\mu(t+1) &= \mu(t) +\gamma(\xi(t)\xi(t)^T\mu(t)-\mu(t)^T\xi(t)\xi(t)^T\mu(t)\mu(t)) + O(\gamma^2)
\end{align}
now, let's put it into continuous domain, so that we can take derivative there, and let's take it with some more realistic assumptions for $\xi(t)$, as following: 
\begin{align}
\dfrac{d}{dt}z(t) = Cz(t) - (z(t)^TCz(t))z(t)
\end{align}
here, $C$ is the covariance matrix of $\xi(t)$. Now we can see that the eigenvectors of $C$ are the solution when we try to maximize the rule updating function. Since $z(t)$ are the eigenvectors of the covariance matrix of $\xi(t)$, we can get the conclusion that $z(t)$ are the principal components of the data. This step can be formally proved with Theorem 2.4 in \cite{hale1969dynamical}, here we don't discuss about it. 

\subsection{Neural Network, Backpropagation, Global Optimum and Universal Approximation }
Here, let's put neurons together and talk about neural networks. 
We will also show its training algorithm, the famous backprogapation and some of its interesting properties 
\footnote{Due to the limit of space, some interesting properties are not covered here, for detailed topics like when neural network is guaranteed to have a global optimal, or when neural network is worse than a simple linear perceptron, please refer to the online version: http://haohanw.blogspot.com/2014/12/ml-my-journal-from-neural-network-to\_15.html}. 
At last, let's have a look at the most famous fact of neural network, its power of universal representation.

\subsubsection{Neural Network}
\begin{figure}[h]
\center
\includegraphics[scale=0.5]{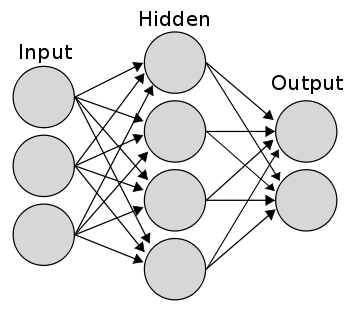}
\caption{A typical two layer neural network. Input layer does not count as the number of layers of a network}
\label{fig:nn}
\end{figure}
When we connect many neurons into a network of neurons, we basically get a neural network. Figure~\ref{fig:nn} shows a typical network, as we can see, more than random connection, the network has a layerwise architecture. 
Input layer takes input data and output layer makes prediction. 

\subsubsection{Backpropagation}
As we know, a single neuron can update its weights during learning through a gradient descent of its error function, backpropagation algorithm is a more generalized situation where we have many more neurons connected layerwisely and the error of a neuron need to be propagated back to the neurons of former layers, unless, of course, current neuron is of the first layer. Neurons of each layer update its weights according to the propagated error.

Let's try to look at this problem formally. We have data $X$ and its golden standard $y$, and a network $N = [l_1, l_2, l_3...]$. 
We first apply the network to the data and get its error as:
\begin{align}
E = \dfrac{1}{2}\sum(N_iX-y)^2
\end{align}
Then calculate the error term for the last layer:
\begin{align}
\delta_{ni}=\dfrac{\partial N_iX -y}{\partial l_{ni}}
\end{align}
Propagate error of layer $n$ to former layer and calculate the error term at layer $n-1$
\begin{align}
\delta_{n-1, i} = \dfrac{\partial\sum_kw_{n-1,i,n,j}\delta_{n,j}}{\partial l_{n-1, i}}
\end{align} 
Then we update the weights according to error term
\begin{align}
w_{n-1,i,n,j} = w_{n-1,i,n,j} + \eta\delta_{n,j}x_{n,j}
\end{align}

Since this report is mainly about deep learning, we don't want to spend too much space for mundane topics for traditional backpropagation. The above derivations should be enough to have an understanding of deep learning. For more information, please refer to the original paper \cite{werbos1990backpropagation}

\subsubsection{Universal Approximation Power}
\label{post:3.4}
Now, let's move to a famous and exciting claim of neural network: a two layer neural network can represent any hypothesis space, it is also known as the universal approximation. 

Before everything, two points need to be made clear. The first one is, as showed in the title, a neural network can only \textbf{approximate} any function, but cannot represent the exact function. 
The second one is that "any function" actually means any continuous function, not functions with sudden jumps.

Although the formal proof of this approximation power is non trivial, we can look at this problem very intuitively. 
For now, let's only consider the linear neurons, which builds the most basic foundation. Neurons with sigmoid squashing output should be easy to understand once the foundation ideas are showed.

\begin{figure}[h]
\center
\includegraphics[scale=0.4]{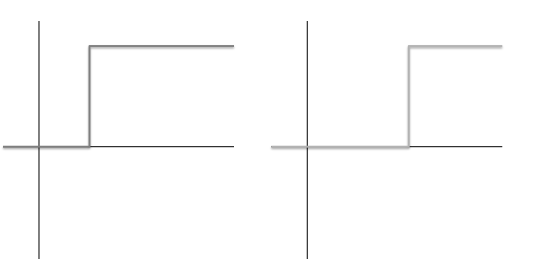}
\caption{Two learnt linear functions with two independent neurons}
\label{fig:uni1}
\end{figure}

Any single linear neuron can be learnt to represent a linear function, as showed in Figure~\ref{fig:uni1}, two learnt linear function from two neurons.

\begin{figure}[h]
\center
\includegraphics[scale=0.4]{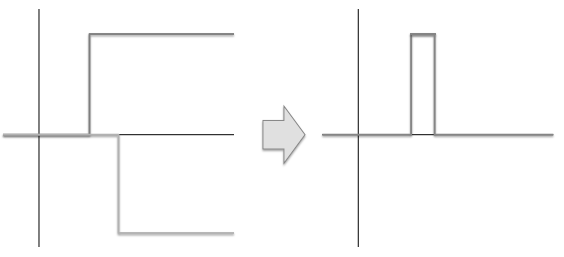}
\caption{A linear combination of two neurons to make a box function}
\label{fig:uni2}
\end{figure}

In Figure~\ref{fig:uni2}, we can see that a linear combination of two neurons can be learnt to represent a box function. Now, we have a box function, of which, the width is controlled by linear combination, and height is controlled by each of neuron. All of these parameters can be learnt through data. Therefore, we can learn any box function with two neurons. Further more, when the width is approximating to 0, we can learn to represent any point with two neurons. Since any function is a set of points, we can approximate any function. This approximation only requires one hidden layer, every pair represents the two basic neuron of a box function, the output layer will combine them linearly.

This theory can be proved more formally \cite{hornik1989multilayer} \cite{castro2000neural}, but we don't cover them here. 

\section{Neural Networks Extended to Deep Discriminative Models}
After some background information in the previous section, now let's start to move to the domain of deep learning. 
In this section, first we need to emphasis the necessity of depth for neural networks, which may sound controversial with previous mentioned theorem of universal approximation power. 
Then we will move to two famous discriminative models: Stacked Autoencoders and Convolutional Neural Networks. 
We will end this section with some optimizing techniques for Convolutional Neural Network, but these techniques can be extended to many other models (even some non-network typed models). 

\subsection{The Necessity of Depth}
In Section~\ref{post:3.4}, we have looked at neural network and talked about its universal approximation property. Then, an interesting question will be, since a neural network is already so powerful, why we need to move to deep learning. 

The take home message here is that, for every one less layer of a neural network, it requires exponential more neurons to approximate the same function. However, the bottleneck of working on a shallow architecture is not simply the expense of computation of exponential more neurons, the requirement of exponential more training data to maintain the generalization ability is a fundamental unsolvable problem. 
Before that, there is also the most intuitive motivation: since one of the most fundamental goals of artificial intelligence is to simulate human intelligence, it's reasonable to simulate the deep brain architecture with learning models (despite the fact that some learning models have already performed better than human). The rest is for a further explanation of these arguments, which is not critical in superficial understanding of deep learning.

\subsubsection{Natural Motivation}
As we have talked, the natural motivation of going into deep architecture is that our human brain forms a deep architecture in recognition.

Although how brain functions still remains as a mystery, people intuitively believe brain forms a hierarchical sensing of knowledge. This belief is validated with some state of art research, like Meunier et al \cite{meunier2009hierarchical}, Perry et al \footnote{http://www.cpri.ca/uploads/section000181/files/brain\%20structure\%20and\%20function.pdf}. We don't have to go into the details of these ideas, since these are not the focuses of a deep learning report. 

\subsubsection{Computation Efficiency}
To understand the computation efficiency of a deep architecture compared to a shallow architecture, let's look at the computation of circuits. 
As we know, a two layer circuit of logic gates can be used to represent any boolean functions, which can be written as either sum of products (disjunctive normal form) or product of sums (conjunctive normal form). 
Further, Hastad \cite{hastad1986almost} has proposed that there are functions computable in polynomial size and depth $k$ but requires exponential size when the depth is restricted to $k-1$.

\begin{figure}[h]
\center
\includegraphics[scale=0.4]{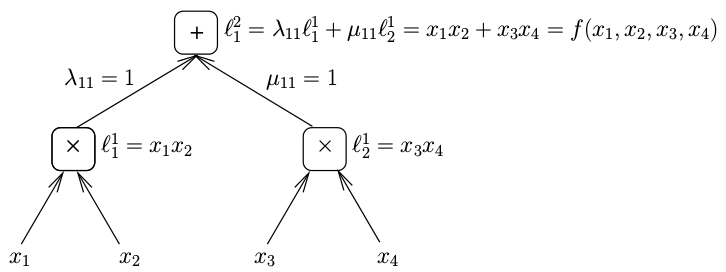}
\caption{A two layer sum-product network. (Figure from the \cite{delalleau2011shallow})}
\label{fig:ds}
\end{figure}

Now, let's move from circuits theory to the networks domain with sum product network. 
A sum product network is an analogy of circuits logic gates in the neural network field, where there are two basic computation neurons. 
A sum neuron is an unit that computes the weighted sum of inputs and a product neuron is an unit that computes the product of inputs.
In \cite{delalleau2011shallow}, Delalleau et al defined a shallow architecture as a sum product network with only one hidden layer, and they defined a deep architecture as a network with more than two hidden layers. 
They introduced a network architecture, where sum layer and product layer stacks alternatively, and each computing neuron only takes in two inputs, as showed in Figure~\ref{fig:ds}.

With these definition, and a series of derivation, they drew the conclusion that the functions that require $O(2^n)$ of units with a shallow network can be represented with $O(n^2)$ units with a depth of $O(log n)$ architecture.

Then, let's move to traditional neural networks. In Section~\ref{post:3.4}, we have seen that any functions can be approximated with a two layer neural network, in which the lower layer generates box functions (or points), the upper layer combines lower layer output. 
This serves as a good analogy of what has been showed in circuits theory. 
Further, a neuron is a linear combination of weights with an activation function. 
By linear combination, a neuron can easily simulate AND and OR relation with a proper set of weights and threshold. \cite{orponen1994computational} proposed a formal theory of it and proved it. 

\subsubsection{Computation vs. Statistics}
Furthermore, when an inefficient representation of a function with shallow architecture is adopted, many more parameters are trained and computed. 
This will easily introduce the curse of dimensionality \cite{friedman1997bias} with insufficient data set. 
Thus, even technology suddenly evolves and engineers find a way to compute exponential number of neurons, the problem of shallow architecture remains as a statistical problem rather than a computational one.

\subsection{Stacked Autoencoders}  
After a long time of preparation, we finally step into the first deep learning architecture. Personally, we think autoencoder is the most simple and intuitive idea of neural networks, so let's start with autoencoder and its deep learning version, stacked autoencoders.

\subsubsection{Autoencoder}
Everything about an autoencoder\cite{liou2014autoencoder} is disclosed by its name. 
"Auto" means an unsupervised learning process, and "encoder" means it learns a coding mechanism (or a representation) of data. 
Like any other coding mechanism, the learnt representation is expected be simpler than data itself. 
Therefore, an autoencoder is a feedforward network that can learn a compressed, distributed representation of data, usually with goal of dimensionality reduction. 
An autoencoder usually has one hidden layer between input and output layer. 
Hidden layer usually has a more compact representation than input and output layers, i.e. hidden layer has fewer units than input or output layer. 
Input and output layer usually have the exact same setting, which allows an autoencoder to be trained unsupervisedly with same data fed into input layer and compared at output layer. 
The training process is the same as traditional neural network with backpropagation, the only difference lies in the error is computed by comparing output to the data itself, rather than supervised labels.

\begin{figure}[h]
\center
\includegraphics[scale=0.8]{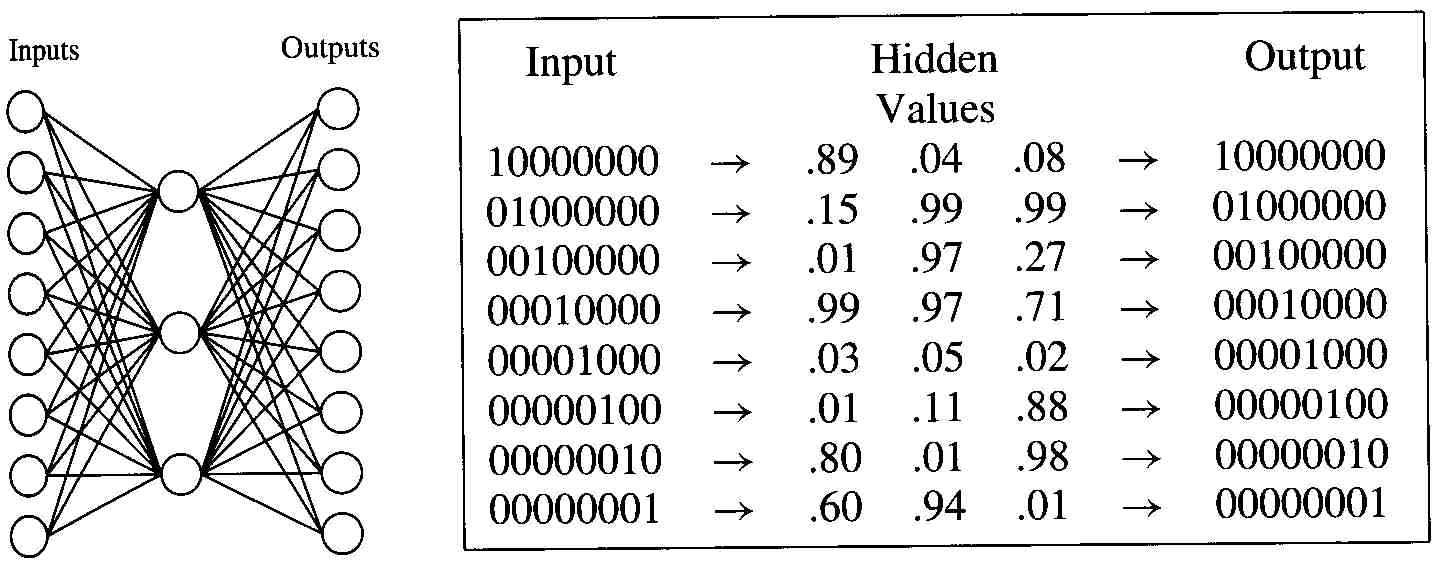}
\caption{Tom Mitchell's example of an autoencoder, (Figure from \cite{mitchell1997machine})}
\label{fig:auto}
\end{figure}

Figure~\ref{fig:auto}. shows my favourite example of an autoencoder, from Tom Mitchell's book \cite{mitchell1997machine}. 
As we can see, the input and output are exactly the same: they are an inefficient representation of eight mutually exclusive labels with bits. 
An autoencoder learns a compact representation with 3 hidden units. 
With a proper threshold applied, the learnt representation is the same as the compact bit representation of eight labels. 
Different from traditional neural networks, what matters in an autoencoder is the representation learnt by hidden layer, rather than the predicted result generated by output layer. 

\subsubsection{Stacked Autoencoders}
As its name discloses, a stacked autoencoders is what we get when we stack a pile of autoencoders together. For every layer, its input is the learnt representation of former layer and it learns a more compact representation of the existing learnt representation.  Thus a stacked autoencoder will have a pyramidal architecture.

\begin{figure}[h]
\center
\includegraphics[scale=0.3]{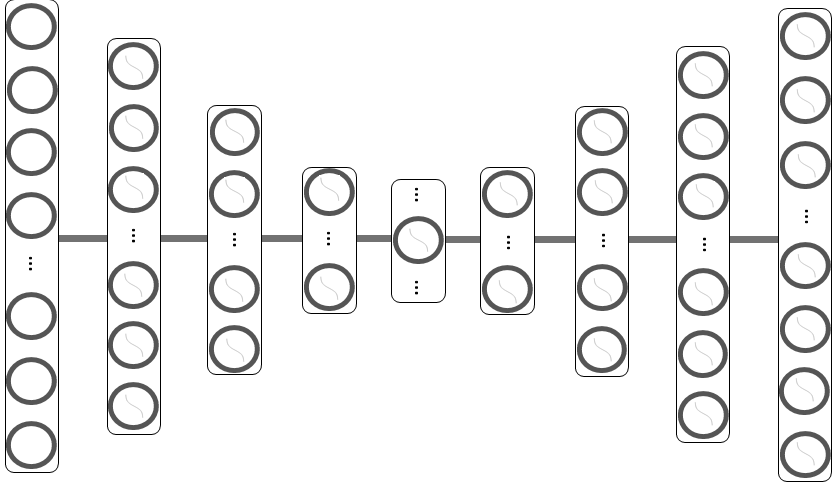}
\caption{A autoencoder with many hidden layers (two stacked autoencoders)}
\label{fig:sda}
\end{figure}

Alternatively, we can think a stacked autoencoders as half of one multi-hidden-layer autoencoder, in which there are $2n-1$ hidden layers, formed with two identical $n$ layer pyramidal architecture connected top layer to top layer. These two top layers merge into one layer since they are identical, thus there are only $2n-1$ hidden layers altogether, as showed in Figure~\ref{fig:sda}.

Since what we are interested in is only the bottom half of what is showed in Figure~\ref{fig:sda}, then how is stacked autoencoders different from any traditional pyramidal structure neural networks? To answer this question, let's first look at a fundamental problem when training a deep neural network with backpropagation. 

\subsubsection{Layerwise Initialization (Pretraining)}
There are several problems when training a deep network with backpropagation algorithm. 
First, when the error is backpropagated, it becomes minuscule in the first several layers and training will not converge to anywhere within a large number of steps. 
This will cause the network to learn to reconstruct the average of all the training data. 
However, this problem could be mitigated when a layerwise learning rate is introduced, allowing lower layers to be trained with smaller learning rates. 
Another problem is that the target space is a non-convex one with many local minimums, with random initialization, backpropagation will often result in an unsatisfying local minimum. 
Thus, the network trained is often useless.

This also explains the reason why deep networks, though apparently able to perform better, cannot outperform shallow architectures until layerwise initialization (pretraining) is introduced in 2006\cite{hinton2006fast}. 
Erhan et al\cite{erhan2010does} have introduced more about the advantages of layerwise pretraining with experiments, mainly for the sake of optimization (as we have discussed) and regularization. 
Layerwise pretraining is very critical in deep learning and we will visit this idea again and again in the future.

This is the reason why we need to think our pyramidal architecture deep neural networks as a stacked autoencoders: we can perform layerwise pretraining. 
For every hidden layer $h_1$, we unsupervisedly train it with its input and an output of identical settings with input. 
Then, we discard the output and stack hidden layer $h_2$ onto it. Where $h_2$ will take $h_1$ as input, and get trained with an output layer with the same setttings of $h_1$. 
Then, again the output layer of $h_2$ is removed, another hidden layer $h_3$ is stacked. 
Same steps repeat until we build enough layers. 

\subsubsection{Stacked Sparse Autoencoders}
Once we have built stacked autoencoders, one choice of next step is to think whether we could learn a compacter representation with existing topology. Some sparsity controls are introduced into the autoencoders to learn a sparse representation. 
Interestingly, with sparsity regularization the architecture of the stacked autoencoders may not necessarily be a pyramid, because regularization will be able to force autoencoders to learn a more concise representation than last layer. 
Thus, although sparse autoencoders are designed to learn a compact representation from a different mechanism, the shared goal (compact representation) of sparse autoencoders and traditional autoencoders leads to an equivalent performance of these two.   

\subsubsection{Stacked Denoising Autoencoders}
As we have seen in Section~\ref{post:2}, an autoencoder trained with squared error criterion is basically a Principle Component Analysis, (when it is trained with other criterion, it is other well defined statistical characteristics of the data set) this is clearly not a desirable solution. We build a deep network to simulate human brain and we expect our deep network could learn some generalization, rather than purely statistics.

In order to train a network with generalization, stacked autoencoders are introduced. 
Its algorithm is as simple as its intuition: human brain is able to discard the noises as it learns a generalized model. 
The algorithm is also very simple, every time when a new layer is trained, the data at input layer is replaced with noised data while the data at output layer stays the same. By doing this, we expect the autoencoders can learn a more generalized and compact representation of data.

\subsection{Convolutional Neural Network}
Here let's have a look at the earliest successful architecture of deep network, the convolutional neural network, also known as LeNet (because of its inventor). Let's look at its basic settings and try to understand some reasoning together with the experimental findings. 

\subsubsection{CNN Architecture}
Inspired by biological understanding of visual cortex, convolutional neural network is proposed based on the foundation of traditional multiple layer perceptrons. 
It is first proposed by LeCun et al in 1998 \cite{lecun1998gradient}, where it got its name LeNet.

\begin{figure}[h]
\center
\includegraphics[scale=0.22]{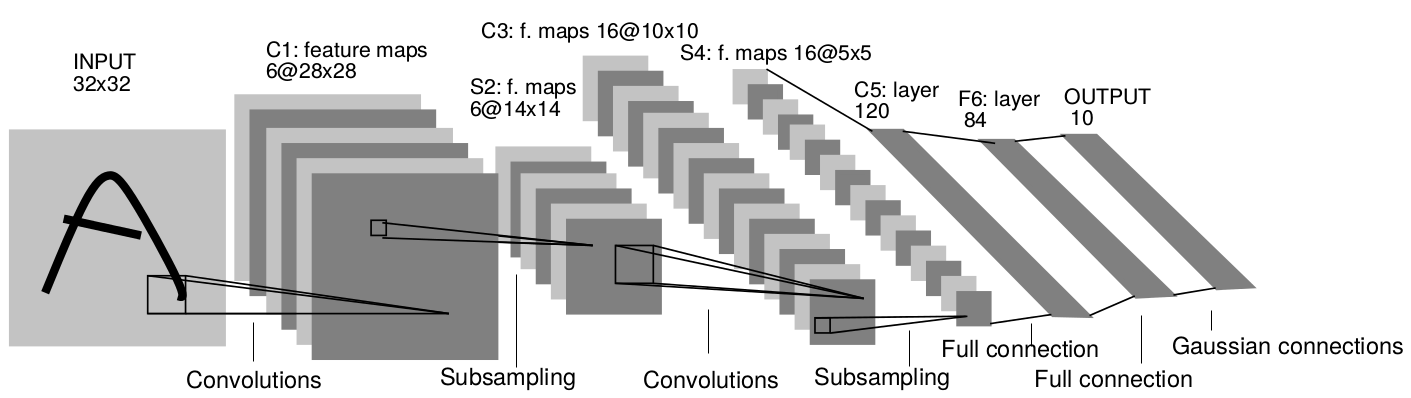}
\caption{Architecture of LeNet-5, one of the first initial architectures of CNN. (Figure from \cite{lecun1998gradient})}
\label{fig:cnn}
\end{figure}

Even from the beginning of CNN, it is very different from traditional neural networks. As researches of all these years go, more and more techniques have been introduced with better test results. Here let's start with the very beginning architecture in the first paper as showed in Figure~\ref{fig:cnn}. 

As showed in Figure~\ref{fig:cnn}, the network first apply convolution operation and subsampling operation alternatively to the input data, with computation units called convolutional layers and subsampling layers respectively. After two groups of such computation, the representation of data in higher layers are fed into a fully connected traditional neural networks, where it finishes its task of a classification problem.

With this architecture, it introduces several advantages compared to traditional neural networks.
\begin{enumerate}
\item Sparsity: CNN is more computation efficient than a traditional neural network. It is sparse in two ways: 
\begin{enumerate}
\item[*] Sparse Connection: A computation unit of layer l is only connected to a certain number of local units of layer l-1, rather than traditional full connection across all the units of two layers. 
\item[*] Sparse Representation: All the units share the same set of weights connecting from the former layer. 
\end{enumerate}
\item Translation invariance: CNN is more robust towards varies of input features. This property also comes in two points of views: 
\begin{enumerate}
\item[*] On convolution layer: convolution is a computation with the property of translation invariance, a network implemented convolution will inherit this property. 
\item[*] On subsampling layer: with the common sampling method: max-pooling, this layer is also translation invariant, because max is a computation that is invariant to rearranges of input features. 
\end{enumerate}
\end{enumerate}

\subsubsection{Convolutional Layer}
In order to understand convolutional layer, let's first have a look at how matrix convolution works and its effect on image processing.

\begin{figure}[h]
\center
\includegraphics[scale=0.5]{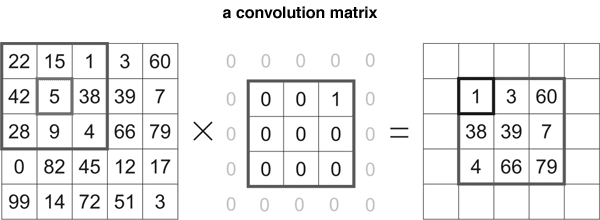}
\caption{An example of matrix convolution}
\label{fig:con}
\end{figure}

We are familiar with the convolution defined on one dimensional functions, it is basically the same when it comes to two dimensional matrices, as showed in Figure~\ref{fig:con}. 

As showed in Figure~\ref{fig:con}, the leftmost matrix is the input matrix, the middle one is usually called kernel matrix. Convolution is applied to these matrices and the result is showed as the rightmost matrix. The convolution process is a element-wise product followed by a sum, as showed in the example. When the left upper 3x3 matrix is convoluted with the kernel, the result is 1. Then we slide the target 3x3 matrix one column right, convoluted with the kernel and get the result 3. We keep sliding and record the results as a matrix. Because the kernel is 3x3, every target matrix is 3x3, thus, every 3x3 matrix is convoluted to one digit and the whole 5x5 matrix is shrunk into 3x3 matrix. (Because 5 - (3-1) = 3. The first 3 is the size of kernel matrix. )

Convolutional layer is a layer that does this job. However, different from traditional designed-kernel image processing, CNN learns the kernel all by backpropagation. The most amazing fact is that, the kernels learnt are usually some meaningful kernel like edge detection designed by human. Using multiple feature maps (computation units) in one convolution layer allows a better combination of kernels to be trained. For instance, in some real world example, a set of edge detection kernel can be trained, with different responsibilities like horizontal edge detection, vertical edge detection.

\subsubsection{Subsampling Layer}
Subsampling is an easier process. It is even easier when we only consider one sampling method for now: max-pooling. It also has a kernel matrix like convolutional layer, but unlike convolutional layer, the computation is as simple as \textit{max}. It is a simple but important computation.

It is responsible for reducing the size of data, thus, reduce computation complexity. Besides, as we have talked above, max is a translation invariant process, it allows CNN to be translational invariant.

\subsubsection{Relevant Techniques}
Now, since we have an idea about CNN, let's move further to see some examples that can make CNN works better.

\textbf{Activation Function} is an important and interesting topic for neural networks. A good selection of an activation function can make a big difference in the performance of a neural network for a certain tasks. Even deep networks are much more sophisticated, this situation remains the same. There is a variety of activation functions, like step function, sigmoid, tanh, softmax, or some complex ones that can capture the desired s-curve shape, or some ones that may not even capture the shape, but somehow work. Some famous examples are as following:\\
1) Hyperbolic tangent function
$$
f(x) = \tanh x = \dfrac{e^x - e^{-x}}{e^x - e^{-x}}
$$
2) Sigmoid function
$$
f(x) = \dfrac{1}{1+e^{-x}}
$$
3) Rectifier function \cite{dahl2013improving}
$$
f(x) = x*I(x>0)
$$
4) Capped Rectifier function. (where $C$ is a constant)
$$
f(x) = \min (x*I(x>0), c)
$$

These functions are usually implemented after one pair of convolutional layer and subsampling layer, take pooling result as input and fire output to next convolutional layer or fully connection layer. Among these functions, the most interesting one for CNN is probably ReLu. 

\textbf{ReLu (Rectified Linear Units)} is an interesting activation function. It allows the network to compute much faster than a network with \textit{sigmoid} or \textit{tanh} as activation functions because it is simply a max operation. More importantly, it allows sparsity of the network since when initialized randomly, around half of the whole network will be 0. Sparsity plays a critical role in deep networks, since it can introduce the advantages like information disentangling, efficient variable-size representation, linear separability etc \cite{glorot2011deep}.

Because of these advantages, ReLu usually plays a role more than just an activation functions integrated into pooling layer. Researchers start to use it as a layer between convolutional layer and subsampling layer, as this demo\footnote{http://cs.stanford.edu/people/karpathy/convnetjs/demo/mnist.html} shows. More than this, some network are built around this idea, like \cite{dahl2013improving}. These works are beyond the domain of this report.

It should be simple to notice that the discontinuity of ReLu at $0$ may hurt the performance of backpropagation. However, \cite{wan2013regularization} has showed that, empirically, ReLu works better than other activation functions like \textit{softmax}, although theoretically it remains an open question.

\textbf{Pooling Function} is used in the subsampling layer to shrink the size of data. Max-pooling comes as the commonest technique probably because it is simple and intuitive. The max of the input features may help the performance of network to the best, but sometime the average of features also works. Therefore, besides max-pooling, there is also one average-pooling.

Besides, \cite{lee2009convolutional} once proposed an idea of probabilistic pooling, where they constrain the pooling process follows a rule that at most of the target unit is on. Thus, the pooling result will be the same as any one of the target unit or it could be nothing. (In comparison, max-pooling always trigger the unit with max value while average-pooling always trigger all the unit and take average of them)

Also, how the pooling area is selected can be a question. \cite{krizhevsky2012imagenet} applies overlapping pooling instead of traditional pooling where the target areas of pooling can overlap each other. 

\textbf{Learning rate} is important in controlling the convergence of gradient based methods. To select an appropriate learning rate is a challenging question. Some smart solutions can be adopted like decreasing learning rates as iteration increases, or training lower layers with greater learning rates and upper layers with smaller learning rates.

\subsection{Dropout \& Maxout}
Now let's look at the details of a famous technique called dropout together with another similar technique called maxout. Again, these techniques are not constrained only to convolutional neural networks, but can be applied to almost any deep networks, or even non-network models.

\subsubsection{Dropout}
\begin{figure}[h]
\center
\includegraphics[scale=0.4]{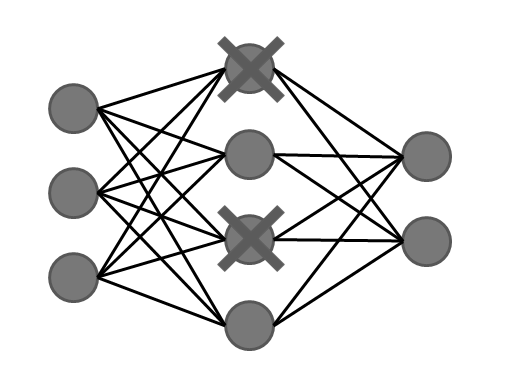}
\caption{An illustration of the idea of dropout}
\label{fig:do}
\end{figure}

Dropout is famous, and powerful, and simple. Despite the fact that dropout is widely used and very efficient, the idea is actually simple: randomly dropping out some of the units while training. One case can be showed as in the Figure~\ref{fig:do}

To state this a little more formally: for each training case, each hidden unit is randomly omitted from the network with a probability of $p$. One thing to notice though, the selected dropout units are different for each training instance, that's why this is more of a training problem, rather than an architecture problem.

As stated in the origin paper \cite{hinton2012improving}, another view to look at dropout makes this it more interesting. Dropout can be seen as an efficient way to perform model averaging across a large number of different neural networks, where overfitting can be avoided with much less cost of computation.

\subsubsection{Maxout}
Maxout is an idea derived for dropout. It is simply an activation function that takes the max of input, but when it works with dropout, it can reinforce the properties dropout has: improve the accuracy of fast model averaging technique and facilitate optimization.

Different from max-pooling, maxout is based on a whole hidden layer that is built upon the layer we are interested in, so it's more like a layerwise activation function. As stated in \cite{goodfellow2013maxout}, with these hidden layers that only consider the max of input, the network remains the same power of universal approximation. The reasoning is not very different from what we did in Section~\ref{post:3.4} on universal approximation power.

Despite of the fact that maxout is an idea derived on dropout and works better, maxout can only be implemented on feedforward neural networks like multi-layer perceptron or convolutional neural networks. In contrast, dropout is a fundamental idea, though simple, that can work for basically any networks. Dropout is more like the idea of bagging, both in the sense of bagging's ability to increase accuracy by model averaging, and in the sense of bagging's widely adoption that can be integrated with almost any machine learning algorithm.

\section{Deep Generative Models}
Now let's visit another family of machine learning models: generative model. In this section, we will first start with an early stage work called self organizing map, which will help us build some understanding of generative models. Then, with a summary of the relationships of generative models we start to visit Hopfield Field, Boltzmann Machine, Restricted Boltzmann Machine and Deep Belief Nets. 

\subsection{Self Organizing Map}
\begin{figure}[h]
\center
\includegraphics[scale=0.6]{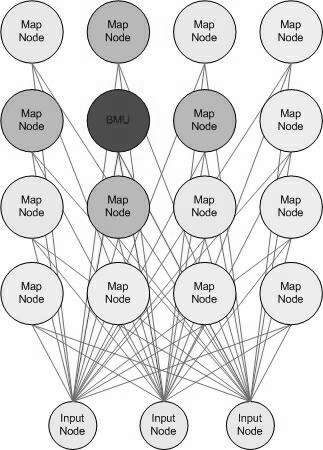}
\caption{ An example of Self Organizing Map}
\label{fig:map}
\end{figure}
Self Organizing Map is a type of neural network invented by Kohonen in 1980s \cite{kohonen1998self} that can be used to represent data with lower dimension, usually with one or two dimensions. It can be trained unsupervisedly, thus the techniques that we are familiar with, like backpropagation or gradient descent is not feasible here. The only thing that remains related to traditional networks is probably how the weights are updates. 
Figure~\ref{fig:map} shows a typical Self Organizing Map in two dimension.

As showed in Figure~\ref{fig:map}, Self Organizing Map is a collection of nodes forming a structure of a lattice. All these nodes are fully connected to the input data. All the input nodes are combined together to represent one input vector, which is of the same length of the weight vector of each of the map node. 

\begin{figure}[h]
\center
\includegraphics[scale=0.12]{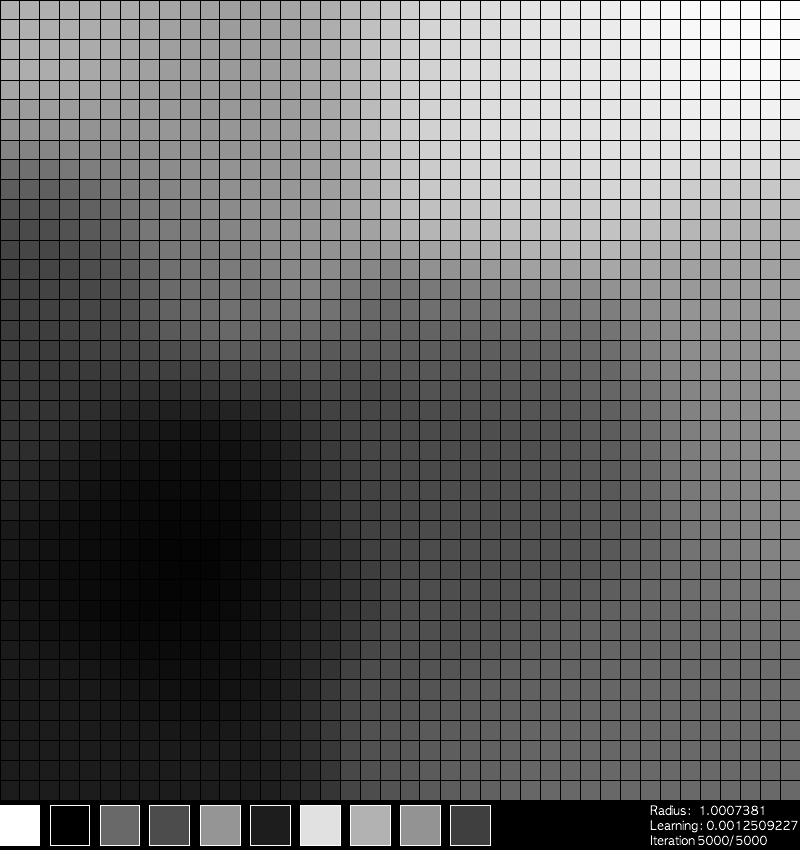}
\caption{An example of Self Organizing Map representing color}
\label{fig:color}
\end{figure}

Now, let's have a look at how a two dimensional SOM can represent higher dimensional data. Figure~\ref{fig:color} is an example when a map is used to represent a 10-dimension color space. As we can see, it looks like a result of clustering, where each color is represented locally.

\begin{figure}[h]
\center
\includegraphics[scale=0.25]{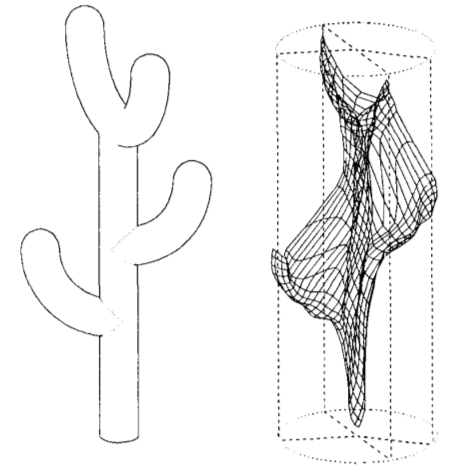}
\caption{A 3 dimensional shape (left) is represented by a two dimensional SOM (right) (Figure from \cite{kohonen1998self})}
\label{fig:shape}
\end{figure}

More interestingly, it works more than representing different data points with a certain area of the map. Figure~\ref{fig:shape} shows how a two dimensional map changes its weights to represent a 3-dimensional shape.

Finally let's conclude the topic of SOM by looking at its training algorithm. The idea is not very complex, but interestingly explains why it could represent data in such a way. As we have said before, the whole idea is not related with backpropagation or so. Some part of it looks like K-means, while some part of it looks like KNN classification. The training algorithm is as following:
\begin{enumerate}
\item First initialization weights of each node randomly
\item When iteration step $t$ does not exceed the maximum allowed steps, pick a random data vector $v$:
\begin{enumerate}
\item[2.1] Select the node with minimum Euclidean distance between node and data, this node is called Best Matching Unit. (BMU)
\item[2.2] Select the nodes of interest as the neighbors of BMU within distance of $r(t)$.
\item[2.3] Update nodes of interest (Node $i$) with the following equation:
\begin{align}
w_i(t+1) = w_i(t) + d_{i, bum}l(t)(v-w_i(t))
\end{align}
\end{enumerate}
\end{enumerate}
where $w$ stands for the weights of these nodes, $d$ stands for a penalty related to the distance between current node and BMU. $l$ is learning rate, which decreases as iteration increases. $r$ is the radius to select nodes of interest, it also decreases as iteration increases.

With this training algorithm, SOM can be trained to represent data with lower dimensions. 

\subsection{Generative (Deep) Network Family}
\begin{figure}[h]
\center
\includegraphics[scale=0.4]{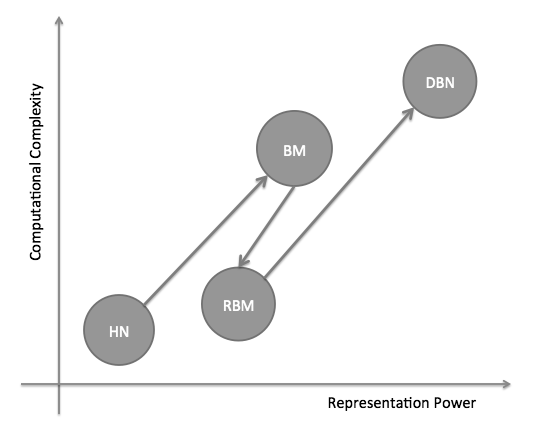}
\caption{An approximation of computation complexity and representation power of each model}
\label{fig:dg}
\end{figure}

Now, we have briefly talked about the very first model in generative model family. Following this, there should come some other famous models, like Boltzmann Machine, Restricted Boltzmann Machine or Deep Belief Nets. Before that, let's first look at some relationship between these three models another one called Hopfield Network. Figure~\ref{fig:dg} shows this relationship. In the figure, x-axis represents the representation power of each model and y-axis represents the computation complexity to perfectly train this model. It only shows this idea approximately. 

In Figure~\ref{fig:dg}, we can see that there is a huge difference in the representation power, as well as computation complexity across all these four models. The following introduction will following the arrows in the figure across these models, when the underlining story is how we step by step, move to deep belief networks. 
First we start from Hopfield Network, which has a very limit representation power, yet it could be trained fast. 
Then researchers want their model to represent more knowledge, so Boltzmann Machine is born as a stochastic version of Hopfield Network. 
However, the problem is that there is no efficient way that can train it. 
Then, researchers start to make constraints, removing all the intra-layer connections lead to Restricted Boltzmann Machine. 
It's more efficient, but the representation is worse as well. 
Then people start to stack RBM and make Deep Belief Nets, which has much more representation power, but also much more difficult to train it perfectly. Fortunately, layer wise initialization makes training feasible and that's why DBN become popular. 
From now on, we will first look at Hopfield Network and Boltzmann Machine, then let's move to Restricted Boltzmann Machine and Deep Belief Nets.

\subsection{Hopfield Network}
Hopfield Network is a recurrent network with binary units, proposed by John Hopfield in 1982 \cite{hopfield1982neural}. In the newtork, recurrent means that all the edges in this network are bidirectional and binary units are the units that can take either one of two values. (Mostly, these two values are 0 and 1, sometimes, it could be -1 and 1. This difference is not just about representation, but also about how the energy is computed, thus about which configuration the network will converge to. As we may see briefly in the following text.)

\begin{figure}[h]
\center
\includegraphics[scale=0.25]{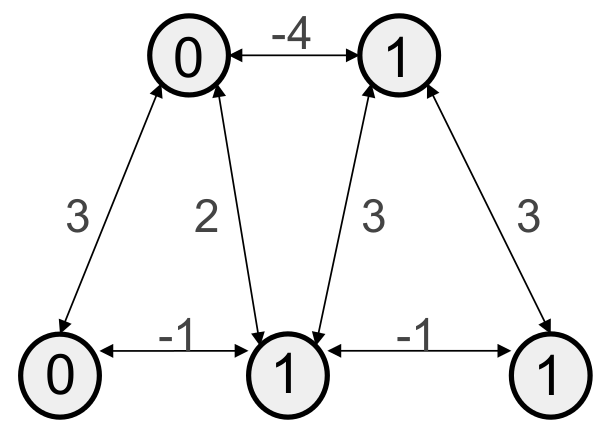}
\caption{An example of Hopfield Network}
\label{fig:hf}
\end{figure}

Figure~\ref{fig:hf} shows a very simple example of Hopfield Network, those numbers on the edge are the weights of the bidirectional edges, numbers in the circles represents the state of the units. Current setting is a stable state of this network. 
The values of weights are predefined, and we will talk about how these stable states of units are achieved after we talked about its energy.

In Hopfield Network, there is no such unit as input unit or output unit. Every unit can be fed in with data and every unit can serve as the input for other units.

\subsubsection{Energy of Hopfield Network}
The energy of Hopfield Network can be defined as following:
\begin{align}
E = \sum_is_ib_i - \dfrac{1}{2}\sum_{i,j}s_is_jw_{i,j}
\end{align}
where $s$ stands for the state of a unit, $w$ stands for weight, $b$ stands for bias, we take half of the second term because it is a symmetric network. 
Here we introduce $b$ to make the equation more formal, but we don't bother with it here.
Therefore, when we ignores the bias term, the energy of the the network in Figure~\ref{fig:hf} is -5. This is the global minimum of energy of this network. 

\subsubsection{Application of Hopfield Network}
The goal of a Hopfield Network is to keep update its units to reach the minimum of energy, which could be used to represent memory. 
This memory has a very interesting application: it allows content addressable memory. 
If a Hopfield Network memorizes an event as the minimum of its energy, then if the event somehow gets corrupted or noised, we could recover all the event by feeding part of that event into this Hopfield Network, when it updates itself to the minimum of energy, those states of unit corresponds to the whole event.

\begin{figure}[h]
\center
\includegraphics[scale=0.9]{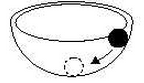}
\caption{An analogy of the memory in Hopfield Network}
\label{fig:bowl}
\end{figure}

As a very interesting analogy, we can think Hopfield Network memorizes the stable state for a ball in a bowl, as showed in Figure~\ref{fig:bowl}

As basic physics tells us, no matter where the ball is placed initially, it will stop at the bottom of the ball stably, where it has the minimum of its energy.

However, Hopfield Network cannot keep the memory very efficient, because a network $N$ units can only store memory up to $0.15N^2$ bits. While a network with $N$ units have $N^2$ edges. Also, after storing $M$ memories, each connection has a integer value in range $[-M, M]$. Thus, the number of bits required to store $N$ units are $N^2log(2M+1)$. Let's not worry about how these numbers are proved, for this report, they are used to show that Hopfield Network is not very efficient to store memory. 

\subsubsection{Learning of Hopfield Network}
Now let's have a look at how Hopfield Network learns the states of its units. It basically follows the Hebbian Learning rule that we have talked about in Section~\ref{post:1}, "Cells that fire together, wire together", since the energy is calculated for each two units and the link between them. This also explains why weights are symmetric. There are two ways to update the unit.

\textbf{Asynchronous} update the unit with a random order, one by one. Each time we select a random unit and to test if switching the unit will result in decreasing the total energy. Then we adjust the unit and move to another unit randomly.

\textbf{Synchronous} update will first test each unit to see whether we should adjust its state according to the current setting. After testing all the units, make the adjustment. This may not even end up with decreasing the energy. 
Both of these two methods may lead to local minimum, oscillation or loop of states. 

\subsection{Boltzmann Machine}

\begin{figure}[h]
\center
\includegraphics[scale=0.6]{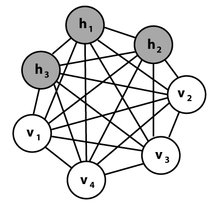}
\caption{An example of Boltzmann Machine}
\label{fig:bz}
\end{figure}

As Figure~\ref{fig:bz} shows, it is a fully connection network with hidden units and visible units, where visible units can be fed in with data. This figure clearly shows why Boltzmann machine is a hidden unit version of Hopfield Network, then how about stochastic?

As we know, in the learning procedure of Hopfield Network, skipping out of local minimum will be tough. So, inspired from physics, a method to transfer state regardless current energy is introduced. For a certain unit, set it to state 1 regardless of current state with the following probability:
\begin{align}
p=\dfrac{1}{1+e^{-\Delta E/T}}
\end{align}

where $T$ stands for the temperature. This is inspired by a physics process where the higher the temperature is, the more likely the state will transfer. Besides, the probability of higher energy state transferring to lower energy state will be always greater than the reverse process. The higher the temperature $T$ is, the less contrast this difference is. This temperature idea is highly related to a very popular optimization algorithm called simulated annealing \cite{kirkpatrick1983optimization} back then, but we don't have to talk about it here. The idea behind simulated annealing is similar to a technique about adjusting learning rate as backpropagation iteration increases. Here, to make this problem simpler, we only worry about the situation when temperature is a constant, for example, when $T=1$. 

\subsubsection{Energy of Boltzmann Machine}
Now let's talked about the energy function of Boltzmann Machine. Similarly to Hopfield Network, the energy of BM is defined as following:
\begin{align}
E(v,h) = -\sum_iv_ib_i - \sum_kh_kb_k - \dfrac{1}{2}\sum_{i,j}v_iv_jw_{i,j} - \dfrac{1}{2}\sum_{i,k}v_ih_kw_{i,k} - \dfrac{1}{2}\sum_{k,l}h_kh_lw_{k,l}
\end{align}
where $v$ stands for visible units, $h$ stands for hidden units, $w$ stands for weights and $b$ stands for bias.

Based on this energy function, the probability of a joint configuration over both visible unit and hidden unit can be defined as following:
\begin{align}
p(v,h)=\dfrac{e^{-E(v,h)}}{\sum_{m,n}e^{-E(m,n)}}
\end{align}
The probability of only visible units (or hidden units) can be achieved by marginalizing this joint probability.

When Boltzmann Machine is trained to its stable state, which is called thermal equilibrium, the distribution of these probability will remain the same because the distribution of energy will remain the same. However, the probability for each visible unit or hidden unit may vary and the energy may not be at their minimum. This is related to how thermal equilibrium \cite{manabe1967thermal} is defined, where the only thing stays the same is the distribution of each part of the system. 

\subsubsection{Application of Boltzmann Machine}
Similar to Hopfield Network, Boltzmann Machine can be used to generate data. By marginalizing the joint probability of visible and hidden units, we can get the probability of visible units as the following function:
\begin{align}
p(v) = \dfrac{\sum_he^{-E(v,h)}}{\sum_{m,n}e^{-E(m,n)}}
\end{align}
when the settings of a BM is learnt, we can easily apply this to the network and sample out the visible units. 

\subsubsection{Learning of Boltzmann Machine}
At last, let's have a look at how to train this powerful Boltzmann Machine by maximizing the likelihood function. Because the probabilities are defined as the exponential term of energy, we try to maximize the sum of log probabilities instead to make it simpler. Now, when we take derivative of the log probabilities over weights, we can see a very beautiful term showed as following:

\begin{align}
\dfrac{\partial \log p(v)}{\partial w} = <s_i,s_j>_{data} - <s_i,s_j>_{model}
\end{align}

It is the difference between the expectation value of product of states while the data is fed into visible states and the expectation of product of states while no data is fed, both products are computed when the system reaches thermal equilibrium. Then the question is about how to get these two products.

The answer is as simple as sampling. Keep sampling all the possible states to until the system reaches its thermal equilibrium. There are several algorithms about this, but we believe the main idea remains the same as sampling. As long as this sampling procedure is not avoided or dramatically simplified, the learning of Boltzmann Machine remains a difficult problem. 

\subsection{Restricted Boltzmann Machine}
\begin{figure}[h]
\center
\includegraphics[scale=0.55]{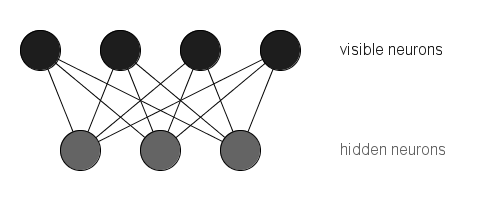}
\caption{An example of Restricted Boltzmann Machine}
\label{fig:rbm}
\end{figure}

Restricted Boltzmann Machine, invented by Smolensky in 1986 \cite{smolensky1986information}, is a Boltzmann Machine without intra-layer connection between units. In other words, there are only connections between visible units and hidden units, but no connections between visible units and visible units or between hidden units and hidden units. It is a bipartite graph, as showed in Figure~\ref{fig:rbm}.

Because of the removal of some connections, the energy function also becomes simpler, as following:
\begin{align}
E(v,h) = -\sum_iv_ib_i - \sum_kh_kb_k - \dfrac{1}{2}\sum_{i,k}v_ih_kw_{i,k}
\end{align}
where, as usual, $v$ stands for visible units, $h$ stands for hidden units, $w$ stands for weights and $b$ stands for bias. 

\subsubsection{Traditional Learning of RBM}
Traditionally, training a RBM follows the same way as training a BM, but it gets much simpler because the removal of intra-layer connections. The adjust of weights is related to the difference of expectation of products of units with data fed in and the expectation of products of without data fed in, as before, as following:
\begin{align}
\dfrac{\partial \log p(v)}{\partial w} = <s_i,s_j>_{data} - <s_i,s_j>_{model}
\end{align}

One simpler thing is that, now we don't have to sample the states one by one to get the thermal equilibrium. Since hidden units do not affect each other now, we can calculate the state of all hidden units within one step parallel. Then we reconstruct visible units, and then compute hidden units. This is Gibbs Sampling \cite{gelfand1990illustration}. However, this only decreases the computation complexity in a sense that all hidden units (or visible units) can be calculated together, but there is not guarantee that how long it takes to get the thermal equilibrium state. 

\subsubsection{Contrastive Divergence}
Contrastive Divergence is an interesting idea, which states that we don't have to keep working until thermal equilibrium. Usually, only a few steps of sampling is enough, sometimes, even one step is enough! (Here one step means first sample hidden states, adjust weights, then reconstruct the visible units, sample hidden units and then adjust weights again.)

This dramatically reduces the computation complexity of training an RBM, but why does this even work? It is only an empirical result when it is first introduced. It could be understood as an changes of error surface, where the first sampling drags down the error surface and the second sampling drags up the sampling to form a local optimum. 

On the theoretical side, Bengio 2007 showed that contrastive divergence (k-step short Gibbs sampling) can obtain a biased but convergence estimator of log-likelihood gradient \cite{bengio2009justifying}. To cover this theory may require heavy work related with Gibbs sampling, so let's skip it here.

One thing to notice: Contrastive Divergence is a biased approximation. It does not maximize the likelihood of the data under the model but the difference of two KL-divergences, like following: 
\begin{align}
KL(q|p)- KL(q_k|p)
\end{align}
where $q$ is the empirical distribution and $p_k$ is the distribution of the visible variables after $k$ steps of the Markov chain. If the chain already reached stationarity it holds $p_k= p$ and thus $KL(p_k|p) = 0$ and the approximation error of Contrastive Divergence vanishes.

\subsection{Deep Belief Networks}
Deep Belief Network is one of the models in the family of belief networks, which is a graph model with directed connections. For a long time, training a belief network is a big problem since there is no effective way to model the posterior probability with explaining away effect. However, in 2006, a breakthrough work\cite{hinton2006fast} which makes training a deep belief network possible. That is the reason that DBN becomes so famous these days. 

\begin{figure}[h]
\center
\includegraphics[scale=0.2]{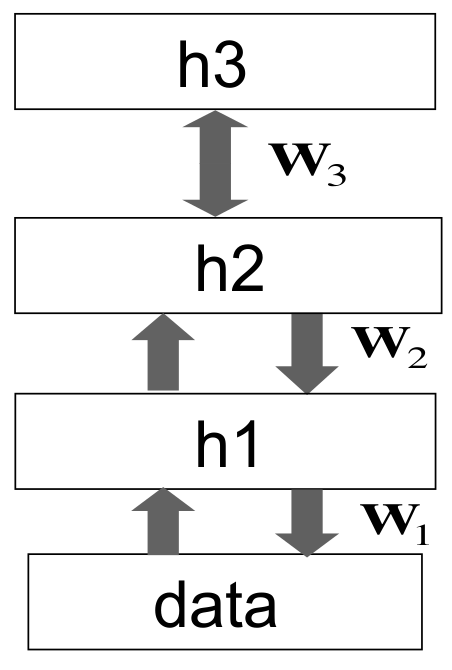}
\caption{A three layer Deep Belief Network}
\label{fig:dbn}
\end{figure}

Deep Belief Network is a stacked RBM. The visible layer of the bottom RBM is fed in with data, and the hidden layer of the bottom RBM is served as the visible layer of the second RBM. Then, a third RBM can be stacked onto the second RBM by merging the visible layer of third RBM and the hidden layer of the second RBM. It could be stacked to as many as possible layers. A 3 layer DBN is showed in Figure~\ref{fig:dbn}.

Same as RBM, a DBN mainly serves as a generative model. When we use it to generate data, we first sample the top layers with Gibbs Sampling to reach its thermal equilibrium, then we pass the visible states of top layer top down to generate the data. This is why in the figure, weights are colored differently. Green arrows are part of the generative models while red arrows are only used for inference.

The breakthrough of how to train DBN is also a simple idea. (But why it works is very complex) The training falls into two steps: the first step is layer-wise pre-training and the second step is fine-tuning. Compared to how we train a neural network traditionally, the first step can also be seen as a clever way of initialization, the second step can be as simple as backpropagation, depends on what model we want to train.

The pre-training step follows the same idea of pre-training a stacked autoencoder. We first treat the bottom two layers as a standard RBM, perform contrastive divergence to train it. Then we move a layer higher, train the second RBM with contrastive divergence, until we finish training the top RBM.

Then, the only question remains is how to fine-tune the model, with different goals of generative model and discriminative model, we have different strategies. 

\subsubsection{Fine-tune for Generative Model}
Fine-tuning for a generative model is achieved with a contrastive version of wake-sleep algorithm. This algorithm is interesting because it is designed to interpret how brain works. Scientists have found that sleeping is a critical process of brain function and it seems to be an inverse version of how we learn when we are awake. The wake-sleep algorithm also has two steps. In wake phase, we propagate information bottom up to adjust top-down weights for reconstructing layer below. Sleep phase is the inverse of wake phase. We propagate the information top down to adjust bottom-up weights for reconstructing layer above.

The contrastive version of this wake-sleep algorithm \cite{hinton1995wake} is that we add one contrastive divergence phase between wake phase and sleep phase. The wake phase only goes up to the visible layer of the top RBM, then we sample the top RBM with contrastive divergence, then a sleep phase starts from the visible layer of top RBM. 

\subsubsection{Fine-tune for Discriminative Model}
The strategy for fine-tuning a DBN as a discriminative model is our old friend: backpropagation. Since we have labels of data, we can apply standard backpropagation to the pre-trained DBN.

However, even we finally apply backpropagation to this network, pre-training is still necessary. By pre-training we can have many advantages like shortening the training time greatly and avoiding poor local optimums. 

\section{Learning Through Time: Modeling Time Series Data}
Now, let's travel back for the one last time. This time, we visit the development track of time series data modelling, or more generally, sequencial data modelling. 
From the beginning of Time Delay Neural Network, we are going to visit famous Recurrent Neural Network and one of this most famous and powerful variant: Long Short Term Memory (LSTM). 
At last, we will see how to build deep recurrent neural networks. 

\subsection{Time Delay Neural Network}
\begin{figure}[h]
\center
\includegraphics[scale=0.3]{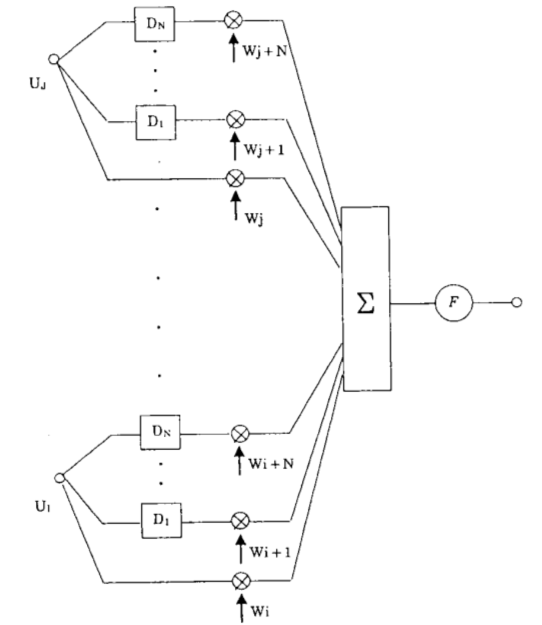}
\caption{A basic local architecture of Time Delay Neural Network (Figure from \cite{waibel1989phoneme})}
\label{fig:tdnn}
\end{figure}

Time Delay Neural Network is invented by Waibel in 1989 \cite{waibel1989phoneme}, it is interesting in the sense that it breaks the limit of traditional neural networks when they only consider the spacial information of data fed to the network. All the data fed in are treated as they occur at the same time and exist for the same period of time by the neural networks, although the features of data may represent different aspects of data of different timing. For example, we can separate a video to a set of frames and concatenate the these frames into one huge vector. This huge vector represents the information of the video with different time point, but when it is fed into a traditional neural network, the neural network only treats them as different information stored in different neurons. This may work sometime, but this is not perfect and that's the reason that TDNN is invented. Figure~\ref{fig:tdnn} shows a typical TDNN, but actual TDNN vary a lot for different applications.

The architecture of Figure~\ref{fig:tdnn} can be seen as a two layer neural network with a complex connections between two layers, or it could also be seen as a single computation unit with only one complicated function that performs summing and thresholding.

As implied in the figure, it is an extension of a traditional neural network which only considers the sum and threshold of an input vector $u$. In a TDNN architecture, several delayed units are introduced. These delayed units are used on keep the information the features ahead. If we are considering features across $t$ steps, then we need $t-1$ delay units. 
The rest of this architecture only varies slightly from traditional neural network. 

\subsection{Recurrent Neural Network}
\begin{figure}[h]
\center
\includegraphics[scale=0.5]{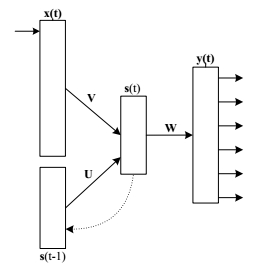}
\caption{A typical example of Recurrent Neural Network}
\label{fig:rnn}
\end{figure}

Recurrent Neural Network\cite{funahashi1993approximation} is a huge family, it is a complement set of traditional neural networks, which we call feed-forward network here. Actually, there should be more recurrent neural networks than feed forward network since RNN removes the constraint that the network only allows information to be passed forward (although another constraint is implied for RNN that there must be at least one feed backward edge in the network). However, in reality, most applications can be done well enough only considering feed forward networks, which leads to the fact that we don't see Recurrent Neural Network so often. However, we are expected to see them a lot in the future because of the success of Deep Belief Networks and Restricted Boltzmann Machines, etc. Basically, every model we have met in the generative deep learning model section are recurrent neural networks since the edges are bi-directional. Let's keep this to the simplest case, as showed in Figure~\ref{fig:rnn}.

As showed in Figure~\ref{fig:rnn}, $x(t)$ stands for input layer, $s(t)$ is a hidden layer, $s(t-1)$ is another hidden layer, $y(t)$ is output layer. Layer $s(t-1)$ gets its input from $s(t)$ and also sends its output to $s(t)$, that this is how this Recurrent Neural Network works. The other parts except $s(t-1)$ works normally as a feed forward network, while $s(t-1)$ stores the information from $s(t)$ and feeds this stored information back to $s(t)$ to be considered as another input at next iteration. As indicated by its name, $s(t-1)$ serves as a unit that can remind $s(t)$ what its last state at last iteration is while $s(t)$ is making decisions at this iteration for the output layer. 

\subsubsection{Backpropagation Through Time}
\begin{figure}[h]
\center
\includegraphics[scale=0.5]{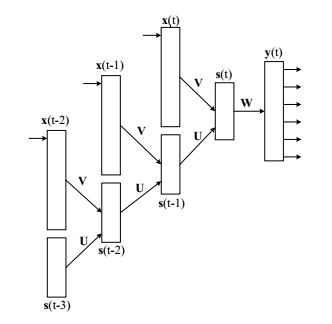}
\caption{An illustration of the unfolding ideas of Backpropagation Through Time}
\label{fig:bptt}
\end{figure}

Traditional backpropagation will not work for recurrent neural networks, for the reason that it requires a clear directed flow of information, so that it knows which direction is "back" to propagate. In RNN, because we have feed back edges, it's not possible to define a back direction. However, Backpropagation Through Time solves this problem. It is a simple extension of backpropagation for RNN. It first applies some modification of recurrent neural network to remove the loops of edges and then applies backpropagation on it. Figure~\ref{fig:bptt} shows a simple illustration on how to modify the network.

As we can see in the figure, BPTT unfolds the recurrent neural network into a feed forward network, where each $s$ layer takes the input from input vectors at current iteration and output of $s$ layer from previous iteration. 
Also shown in the figure, all the weights $U$ share the same parameters because it is originally the same connection, the same applies weights $V$.

After BPTT unfolds RNN to a feed forward network, the only thing left is to apply backpropagation to this network. The only difference is that weights $U$ and $V$ are adjusted according to an aggregated differences of difference of each copy of $U$ or $V$.

One nature question could be, how far into history this BPTT should unfold the recurrent network. In the figure, it only unfolds back to 3 iterations ago. The trade off of selecting a proper length of history to consider should be obvious, the longer the history the model considers, the better the modelling usually is, but the more computation we need to afford and more complex error surface we are working on. In fact, recurrent neural networks already works on quite complex error surface to optimize. Gradient based solution applied here may suffer from more problems of local optimal than applied to feed forward neural network. 

\subsection{Long Short Term Memory}
A typical recurrent neural networks may have a problem remembering inputs several iterations ahead, which means that the influence of a prior input may vanish as the followed up input mixes into the unit and washes out prior influence. Thus the influence of the prior input decays exponentially and this is not a desired property. To solve this problem, Long Short Term Memory recurrent neural network is proposed by Hochreiter et al in 1997 \cite{hochreiter1997long}.

\begin{figure}[h]
\center
\includegraphics[scale=0.2]{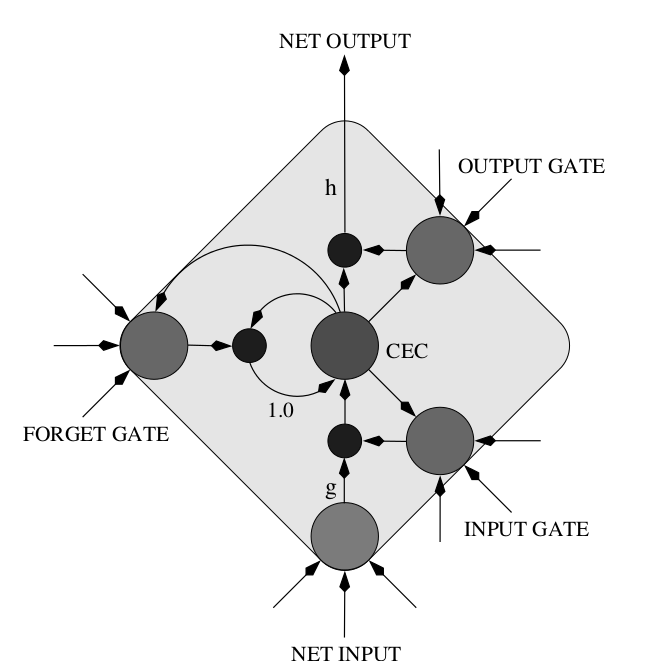}
\caption{A single computation unit for LSTM}
\label{fig:lstm}
\end{figure}

A unit of LSTM is designed to be very complex, as showed in Figure~\ref{fig:lstm}.

The three \textit{GATE} circles are three gates to control the behaviour of LSTM: it only takes new input when the input gate is open, only sends out output when output gate is open and it only keeps remembering its state when the forget gate is open. The middle circle is used for remembering its current state, with a self-directed edge, given that forget gate is open.

With this architecture, LSTM solves the vanishing gradient problem and has been successfully applied to many real world problems. 

\subsubsection{Training of LSTM}
Training of LSTM also follows the standard procedure of recurrent neural networks, which is Backpropagation Through Time. However, it has an advantage over traditional recurrent neural networks under this algorithm, which is that it can preserves the gradient error as it preserves input, such that vanishing gradient\cite{hochreiter1991untersuchungen} problem is avoided.

\subsection{Deep Recurrent Neural Networks}
\begin{figure}[h]
\center
\includegraphics[scale=0.2]{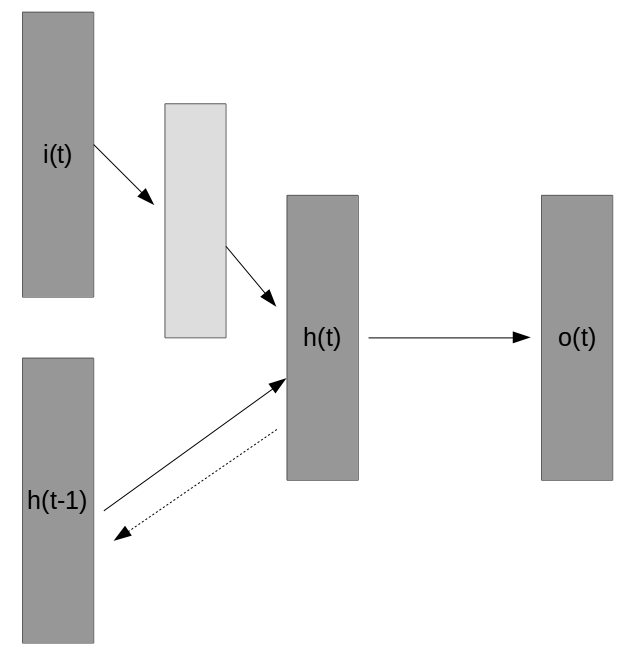}
\caption{Recurrent neural network with deep input architecture}
\label{fig:rnn1}
\end{figure}

\begin{figure}[h]
\center
\includegraphics[scale=0.2]{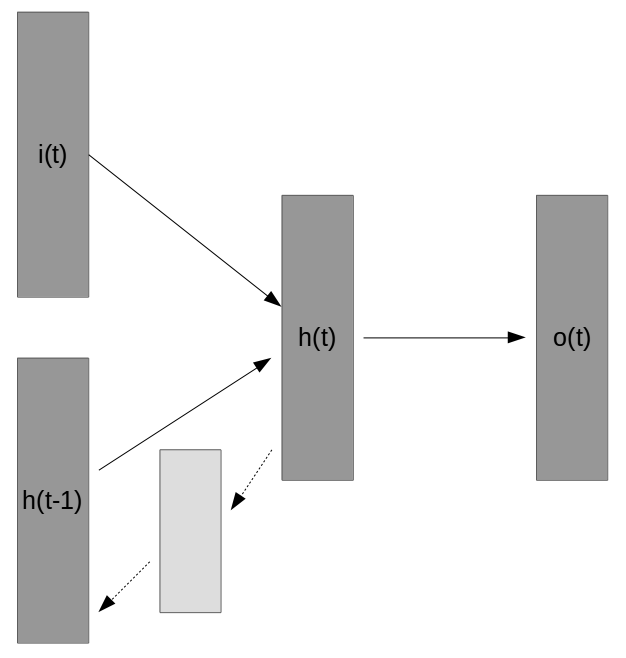}
\caption{Recurrent neural network with deep hidden architecture}
\label{fig:rnn2}
\end{figure}

\begin{figure}[h]
\center
\includegraphics[scale=0.2]{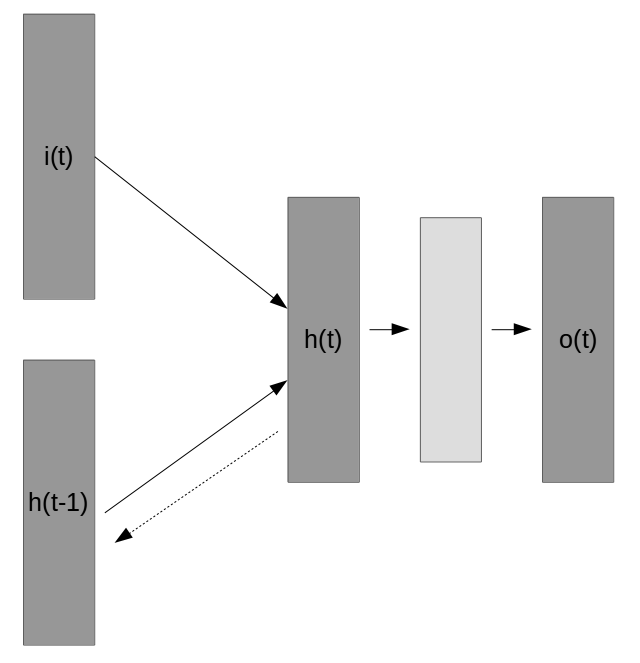}
\caption{Recurrent neural network with a deep output architecture}
\label{fig:rnn3}
\end{figure}
As we have talked previously, since there is no clear direction defined for recurrent neural network, it's also hard to define a deep recurrent neural network itself. However, since a recurrent neural network is consist of three main components: input, hidden layer, output, as well as three main connections: input-hidden, hidden-hidden, hidden-output, it's not hard to expand these basic components with the idea of deep learning, as Pascanu et al did \cite{pascanu2013construct}. We can see these three different deep recurrent neural networks in Figure~\ref{fig:rnn1} to Figure~\ref{fig:rnn3}. In those figures, blue layers are the original layers that a traditional recurrent neural network will have, grey layers are inserted to allow more abstract representation of certain knowledge, i.e. the layers that account for "deep". In the figure, $i(t)$ stands for input, $h(t)$ stands for hidden layer, $o(t)$ stands for output layer.

Figure~\ref{fig:rnn1} shows the simplest version of a deep recurrent neural network. By adding more layers on the input side, a deep RNN allows the hidden layer to learn a more abstract representation of input knowledge. However, this is simple because this is not directly related to recurrent neural network. This representation can be trained and learnt thorough any other deep architecture, like stacked autoencoders, deep belief networks, or even convolutional neural networks, independently from recurrent neural networks.

However, not every deep local architecture can be separated from recurrent neural network. Figure~\ref{fig:rnn2} shows an architecture that learns a better representation of hidden states, that it only works when effectively fed together with the next input.

The last basic combination is deep output layer as showed in the Figure~\ref{fig:rnn3}. Usually we can think this as a multi-layer MLP, but deep generative models can also be used here.

So, these are three basic components for a deep recurrent neural networks. As we can see, they are all simple ideas derived from basic deep networks. A deep recurrent neural network can of course be an arbitrary combination of these three components. A cleverer combination can also be applied, like a combination of a deep and shallow architecture, resulting a "bypass" connection passing information across layers. 

\section{Advantages and Disadvantages of Deep Learning}

\begin{figure}[h]
\center
\includegraphics[scale=0.4]{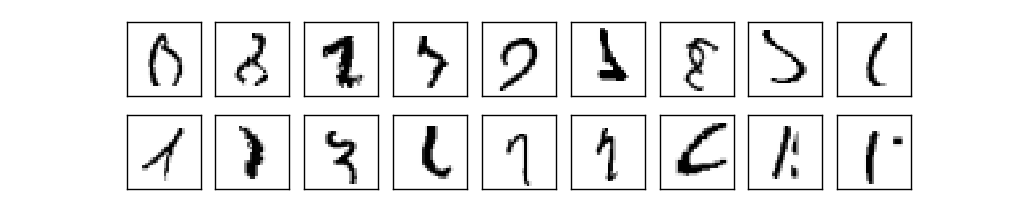}
\caption{Some examples that are impossible to be recognized even by human in MNIST data set. }
\label{fig:mnist}
\end{figure}

Now, let's stop travelling back into the history and start to look forward into the future of deep learning. We all know that it's the most popular machine learning domain so far, but how far it could reach? On one hand, some researcher believe that some traditional problems can be settled by deep learning once for all, for example, deep learning can almost recognizing MNIST data set as well as human being recognition. Figure~\ref{fig:mnist} shows that it basically not possible for machine learning models to reach $100\%$ of recognition accuracy on MNIST data set.  
On the other hand,  Szegedy et al\cite{szegedy2013intriguing} found some intriguing flaws of deep learning, that may lead to a disaster of this domain.

In this final section, let's first talk about the bright side of deep learning with a comparison of the last famous algorithm: SVM, and then talk about some of its flaws. 

\subsection{Theoretically Best So Far}
Let's first look at a Hamlet story of machine learning world. Back in 1980s or early 1990s, Neural Network is like the king of machine learning domain. With a universal approximation power, as we talked in Section~\ref{post:3.4}, and its intuitive advantage of simulating human brain, it is the favorite machine learning algorithm of researchers across every domain. All kinds of extensions have been proposed for different specific problems. However, the crown does not stay long. SVM, invented in late 1990s, soon replaced it. A plain SVM can works as well as a shallow neural network, and proper designed kernelized SVM can also serve as a universal approximation algorithm, but SVM works much faster than neural network. Then, around ten years later, deep learning comes back, as an advanced version of neural network to challenge the leading place of SVM. Basically, it could deprive SVM of leading position easily for the following reason.

According to Yann LeCun\footnote{http://www.kdnuggets.com/2014/02/exclusive-yann-lecun-deep-learning-facebook-ai-lab.html}
, SVM serves like a two layer neural network, where the first layer are support units, measuring the distance between input vector and support vectors and second layer serves as a linear combination of these distance. These distances are not limited to Euclidean, but any similarities that can be calculated with kernels. Yann believes that, deep learning can beat SVM easily simply because that deep learning has more layers, despite the fact that SVM usually searches for global optimum.

Caruana et al\cite{caruana2006empirical} has shown that, among all the popular machine learning algorithms, SVM and Random Forest are generally the best. Since deep learning is more promising than SVM, as we have talked, then how about Random Forest? Again, let's answer this question by understanding algorithm with layers.

\cite{bengio2009learning} has pointed that Random Forest has only one more layer than Decision Tree. A decision can be seen as a two layer classifier, of which the first layer is to match input vector to training data (leaves of a tree) and the second data serving as a linear combination of this exact match. Different of SVM, the second layer serves as a linear combination of exact match, so a decision tree has no power of generalizing knowledge to functions that are not represented by training data. Random Forest builds one more layer on its decision trees, thus it allows the classifier represent exponential more functions. The representation power of a Random Forest is still limited within 3 layers, though there seems no empirical proof to show that this 3-layer architecture is not good enough to compete deep learning. 

\subsection{Some Critical Flaws}
\begin{figure}[h]
\center
\includegraphics[scale=0.9]{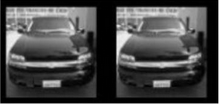}
\caption{Two objects that can be recognized differently by deep learning architecture.}
\label{fig:flaw}
\end{figure}

Interestingly, while people keep discussing about how good a deep learning can be, some experimental results surprise human beings about how stupid it could be. Let's look at Figure~\ref{fig:flaw}. How many cars are there in it?

Obviously, there are two.  Interestingly, according to Szegedy et al's finding \cite{szegedy2013intriguing}, a deep learning architecture will believe that there is only one car because it thinks the left one is a car while the right one is something else, a truck, a horse, or even an orange.

To clarify, these two pictures are not exactly the same, there are little variations in many pixels of these images, but these variations are not noticeable by human eyes.

It's already surprising enough that a deep learning cannot recognize these two pictures into same class. It is more surprising that this is a universal problem. Almost every deep network has these "blind spots". For every deep network, one can manually manipulate some pixels of a figure to make the network recognize this figure into something else.

\begin{figure}[h]
\center
\includegraphics[scale=0.35]{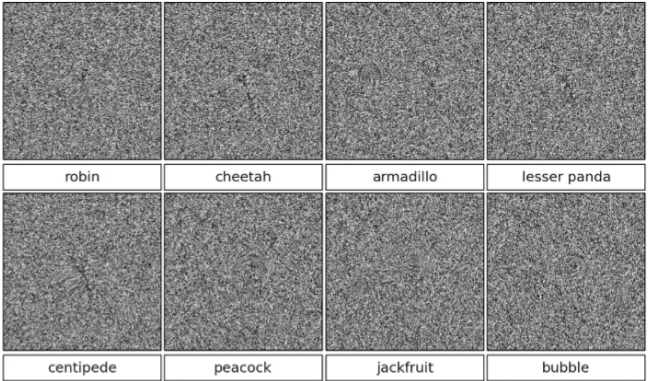}
\caption{Some human generated figures that are recognized as familiar objects with high confidence ($>99.6\%$)}
\label{fig:flaw2}
\end{figure}

Another interesting property that makes the power of deep learning controversial is raised by Nguyen et al \cite{nguyen2014deep}. 
They have showed that a state-of-art deep learning model trained by ImageNet, that can perform well on normal tasks, recognize non-recognizable images as familiar objects with a very high confidence. 
Figure~\ref{fig:flaw2} shows some of such human generated noised images are recognized by deep neural networks. 

This example shows us that although deep neural networks are generally believed to be the models that have the most promising chance to simulate human brain, there will be still a long way before neural networks can play the role human brain plays, despite the fact that these models behave quite well on certain tasks. 

\section{Conclusion}
In this report, we have traced back to the very beginning of deep learning research and have reviewed how those deep learning models have evolved from the initial ideas to the powerful model of today.

Our road map follows the trace that, we start from how people start to pay attention to human brain and model brain as a connection of neurons. Then, this connection of neurons model is extended to nowadays neural network. With some knowledge of human visual system, convolutional neural network is developed as an extension of neural network, and as the pioneer member of deep learning family.

On the othe hand, Self Organizing Map is invented for representation learning. Following the generative model's track, Hopfield Network leads to Boltzmann Machine, then leads to Restricted Boltzmann Machine. Finally RBM is stacked up to be the famous Deep Belief Nets these days.

Following the track of modelling sequential data, there are Time Delay Neural Network, Recurrent Neural Network and famous Long Short Term Memory. This report also reviewed deep recurrent neural network, which is a intuitive combination of deep networks and RNN.

Besides these history of evolution of different deep learning models, it also showed some interesting facts and future challenges of deep learning models. This report can help to understand of how these models are developed and help to contribute in developing new models. 

\section*{Acknowledgements}
Thanks to the Professor Bhiksha Raj for the instruction during his lab course and thanks to the teaching assistant of that course, Zhengzhong Danny Lan\footnote{http://www.cs.cmu.edu/~lanzhzh/}. 

\bibliographystyle{acm}
\bibliography{report}

\end{document}